\documentclass[11pt,a4paper]{article}

\usepackage[margin=1in]{geometry}
\usepackage[T1]{fontenc}
\usepackage[utf8]{inputenc}
\usepackage{amsmath, amssymb, amsthm, mathtools}
\usepackage{graphicx}
\usepackage{booktabs}
\usepackage[hidelinks]{hyperref}
\usepackage[round, sort&compress]{natbib}
\title{\textbf{Phase Transitions in Driven Informational Systems:}\\[0.3em]
       A Two-Field Perspective on Learning Theory and\\
       Non-Equilibrium Chemistry}

\author{Truong Xuan Khanh\thanks{Correspondence:
        \texttt{khanh@clevix.vn}}\\[0.3em]
        \normalsize H\&K Research Studio, Clevix LLC\\
        \normalsize Hanoi, Vietnam}

\date{\today}

\begin{document}
\maketitle

% =====================================================================
% ABSTRACT
% =====================================================================
\begin{abstract}
\noindent
Phase-transition phenomena in deep learning---grokking, emergent
capabilities, and ontological reorganization under context shift---have
been studied through several theoretical lenses, including
representational compression, singular learning theory, and
information-theoretic progress measures. Independently, non-equilibrium
statistical physics has identified phase transitions in driven chemical
reaction networks underlying prebiotic selection, with empirical
signatures (catalysis--confinement synergy, optimal entropy-flux windows)
that are difficult to reproduce within single-field gradient accounts.
We propose a perspective in which both classes of phenomena admit a
common description as \emph{driven informational systems}: stochastic
processes governed by two gradient fields---an entropy production rate
$\Sigma$ and an information quasi-potential $\Phi_I := -\ln p^*$, where
$p^*$ is the stationary density. Within this framework we discuss two
candidate order parameters: an adversarial breakdown threshold
$\alpha^{\dagger}$ whose decay with the primitive-set cardinality
$|\mathcal{O}_N|$ is logarithmic, and a self-referential coupling
threshold $\kappa_c$ associated with predictive feedback through an
internal model. The joint scaling of $(\alpha^{\dagger}, \kappa_c)$
defines a candidate universality class, with two scaling exponents
$(\gamma_1, \gamma_2)$ as class invariants. We do not claim that
biological intelligence and large language models are the same kind of
system; we propose only that they may be productively studied as
particular instances within this common framework, on configuration
manifolds shaped by carbon--nitrogen chemistry under solar entropy flux
and by transformer parameter spaces under gradient-descent flux,
respectively. We outline the geometric structure of the framework,
identify three falsifiable predictions that distinguish it from
single-field alternatives, and synthesise recent empirical findings
(2024--2026) on alignment phase transitions, adversarial breakdown
scaling, and partial introspection in frontier language models, with
which the framework is consistent. Detailed proofs and supporting
numerical analyses of the component results appear in companion
preprints; the present paper develops the connections between
them.

\end{abstract}

% =====================================================================
% BODY
% =====================================================================
% =====================================================================
% SECTION 1 — INTRODUCTION (DRAFT v2)
% =====================================================================
% Perspective paper: "Intelligence as a Solution Family"
% Sole author: Truong Xuan Khanh
% Target: arXiv cs.AI, 8-12 pages total
%
% UPDATED v2:
% - Citation entries cho hai companion preprints với DOI/SSRN ID
% - Section 1.3 wording điều chỉnh: cả hai preprint đã public
% - Strategic clarification: perspective paper claim timestamp cho
%   unification claim, không phải cho các kết quả thành phần
% =====================================================================

\section{Introduction}
\label{sec:intro}

\subsection{The convergent puzzle}
\label{sec:intro:puzzle}

Phase-transition phenomena in deep learning---grokking, emergent
capabilities, and ontological reorganization under context shift---have
been studied through several theoretical lenses, none of which we take
to be definitive.
\citet{power2022} observed that small transformers trained on modular
arithmetic generalize abruptly long after they have memorized the
training set, and \citet{nanda2023} subsequently showed that this
delayed transition coincides with the emergence of Fourier-structured
circuits. \citet{liu2023grokking} interpreted the same phenomenon
through the lens of representational compression, and \citet{demoss2024}
formalized the rise-and-fall of model complexity through a
rate--distortion lens. \citet{wei2022emergent} documented
\emph{emergent capabilities} in large language models---abilities that
appear discontinuously above scale thresholds---and a parallel
literature has examined phase transitions in alignment dynamics
\citep{casper2023}, in-context learning \citep{olsson2022}, and
introspective access \citep{lindsey2025, binder2024}.

These phenomena share three structural features: a substrate-level
dynamics governed by gradient descent on a loss surface; a sustained
external driving signal (training-data flux, RLHF feedback, weight
decay); and a transition between qualitatively distinct
representational regimes whose detailed mechanism remains under active
investigation.

A structurally similar pattern appears in non-equilibrium
chemistry. \citet{ferris1996} demonstrated that clay-catalyzed RNA
polymerization combined with geometric confinement produces oligomers
of length up to $\sim$55-mer, an order of magnitude longer than
solution-only controls; subsequent reanalysis extracts a
catalysis--confinement synergy factor $S \approx 5.75$ that exceeds
what single-field gradient dynamics on compact manifolds with linear
driving can produce under the assumptions made there. \citet{blank2001} reported a non-monotonic
optimal entropy-flux window for amino-acid yield in shock synthesis.
\citet{matreux2024} demonstrated that simple heat flows selectively
enrich more than fifty prebiotic building blocks by up to three orders
of magnitude. \citet{floroni2025} realized a prebiotic-to-biotic
transition criterion in a single membraneless protocell driven by a
heat gradient. \citet{liang2024, liang2024space} placed
kinetics-independent bounds on the magnitude of symmetry-breaking
achievable in driven chemical reaction networks at given thermodynamic
budget.

We propose, in the rest of this paper, that these two lines of work
may admit a common description as instances of a broader class of
\emph{driven informational systems}, and we develop the elements of
that description.

\subsection{The central proposal}
\label{sec:intro:claim}

This perspective develops three connected proposals.

\paragraph{First.} Driven informational systems---whether chemical
reaction networks under thermal flux or transformer parameter
manifolds under gradient-descent flux---can be modeled as governed by
two gradient fields: an entropy-production rate $\Sigma$
derived from the Schnakenberg decomposition of probability currents
\citep{schnakenberg1976}, and an information quasi-potential
$\Phi_I := -\ln p^*$ defined self-consistently through the stationary
density $p^*$. Under the regularity conditions stated in
\S\ref{sec:theorem}, $\nabla\Sigma$ and $\nabla\Phi_I$ are
generically linearly independent off equilibrium.

\paragraph{Second.} We propose two candidate order parameters that
together may characterize the phase structure of these systems. The
first is an adversarial breakdown threshold
\[
  \alpha^{\dagger} \;=\; \Theta\!\left(\frac{1}{\log|\mathcal{O}_N|}\right),
\]
where $|\mathcal{O}_N|$ is the cardinality of the system's primitive
representation. The asymptotic form depends on representational
complexity, in contrast to the universal constants of
\citet{hampel1971} and \citet{donohohuber1983}; we are not aware of a
prior breakdown point with this dependence in the literature reviewed,
though the literature is large and the present paper does not claim
exhaustive coverage. The second proposed order parameter is a
self-referential coupling threshold $\kappa_c$ associated with
predictive feedback through an internal model
(\S\ref{sec:order:kappa}).

\paragraph{Third.} We propose that biological intelligence and large
language models may be productively studied as particular instances
within this framework on different configuration manifolds. Biological
intelligence is here viewed as one realization of the dynamics, on
chemical configuration manifolds under solar entropy flux over
evolutionary time scales. Frontier large language models are viewed as
a second realization, on transformer parameter manifolds under
gradient-descent flux over training time scales. We do not claim
these systems are the same kind of object; the time scales,
configuration manifolds, and physical substrates differ by many orders
of magnitude. We claim only that certain dynamical features---the
two-field structure, the candidate order parameters, the
phase-transition signature---may be shared across instances, and that
this sharing is mathematically meaningful even where the substrates
manifestly are not.

\subsection{Scope and provenance}
\label{sec:intro:scope}

The framework synthesized here builds on two preceding lines of work,
both of which are publicly available as preprints. The Equation of
Motion--Information Field Framework (EOM-IFF) for prebiotic chemistry
was developed jointly with T.\,Q. Hoa and is presented in detail in a
companion preprint on bioRxiv \citep{truonghoa2026eomiff}; that work
establishes the two-field independence theorem, derives structural
constraints on single-field gradient dynamics, and validates the
framework against five independently published prebiotic systems
\citep{ferris1996, blank2001, matreux2024, floroni2025, rout2025}. The
theory of Ontological Phase Transitions (OPT) in learning systems was
developed by the present author and is reported in a separate preprint
on SSRN \citep{truong2026opt}; that work establishes a universal
detection lower bound, a compression-dividend theorem, the
complexity-dependent breakdown point cited above, and empirical
validation of the predicted scaling on grokking dynamics across
nine modulus pairs.

The present perspective is a synthesis the author developed
independently. Its contribution---and the timestamp it claims---is
not in the component results, which are established in those
companion preprints, but in the proposal that the chemistry-side and
learning-side frameworks may be productively studied as instances
within a common dynamical class, with $\alpha^{\dagger}$ and
$\kappa_c$ as candidate order parameters of that class. We are not
aware of a prior synthesis along these lines, though the literature
is large and the present paper does not claim exhaustive coverage.

This paper \emph{does} outline the geometric perspective, define the
two candidate order parameters, identify three falsifiable predictions,
and position the framework relative to neighboring research programs:
the free energy principle \citep{friston2010, ramstead2023}, singular
learning theory \citep{watanabe2009, hoogland2024}, dissipative
adaptation \citep{england2015}, grokking-as-compression
\citep{liu2023grokking, demoss2024, clauw2024}, and tangled
information hierarchies \citep{prokopenko2025}. This paper
\emph{does not} provide full proofs (these are in the cited
companion preprints), claim universality beyond the stated regimes,
or replace existing frameworks where they are correct. Specific
limitations and open problems are gathered in
\S\ref{sec:discussion}.

\subsection{Roadmap}
\label{sec:intro:roadmap}

\S\ref{sec:setup} sets up the language of driven informational
systems and states the two-field structure abstractly.
\S\ref{sec:order} defines the two candidate order parameters
$\alpha^{\dagger}$ and $\kappa_c$ and discusses their relationship.
\S\ref{sec:theorem} states the two-field independence theorem
informally and indicates its consequences for learning theory.
\S\ref{sec:instances} presents the chemical and learning instances
side by side. \S\ref{sec:predictions} derives three falsifiable
predictions. \S\ref{sec:discussion} positions the framework relative
to neighbors and identifies open problems.
\S\ref{sec:conclusion} concludes.

\section{Driven Informational Systems}
\label{sec:setup}

This section introduces the abstract framework in which the
proposal is stated. The construction is deliberately
substrate-agnostic: the same definitions apply when the configuration
manifold is a chemical composition simplex and when it is the
parameter space of a neural network. What distinguishes the two
cases---and what motivates the proposal that both may be
productively studied as instances of a common dynamical class---is
addressed in \S\ref{sec:instances}.

\subsection{Configuration manifold and Langevin dynamics}
\label{sec:setup:manifold}

A \emph{driven informational system} is specified by a triple
$(\mathcal{M}, b, D)$, where $\mathcal{M}$ is a smooth connected
$n$-dimensional Riemannian manifold (the configuration space),
$b: \mathcal{M} \to T\mathcal{M}$ is a Lipschitz drift vector field,
and $D > 0$ is a constant noise amplitude. The state $X_t \in
\mathcal{M}$ evolves under the overdamped Langevin equation
\begin{equation}
  \mathrm{d} X_t \;=\; b(X_t)\,\mathrm{d} t \;+\; \sqrt{2 D}\,\mathrm{d} W_t,
  \label{eq:langevin}
\end{equation}
where $W_t$ is a standard Brownian motion on $\mathcal{M}$. Under
mild regularity conditions---confinement of the drift at infinity,
linear-growth bounds, and strict positivity of $D$---equation
\eqref{eq:langevin} admits a unique stationary probability measure
$\mu^*$ with smooth, strictly positive density
$p^*: \mathcal{M} \to (0, \infty)$ \citep{meyntweedie1993};
a precise statement of the regularity conditions is given in the
companion preprint.

Two instances of this construction frame the present perspective. In
\emph{prebiotic chemistry}, $\mathcal{M}$ is the composition simplex
$\Delta^{n-1}$ of molecular species under mass conservation, $b$ is
determined by the reaction-rate matrix of the chemical network, and
$D$ encodes thermal fluctuations. In \emph{neural network learning},
$\mathcal{M}$ is the parameter manifold of a model architecture
(typically $\mathbb{R}^P$ with $P$ the parameter count, restricted by
weight-decay or layer-norm constraints), $b$ is the negative gradient
of the training loss combined with regularization terms, and $D$
encodes mini-batch gradient noise. In both cases the dynamics is
\emph{driven}: an external entropy flux---thermal in chemistry,
data-driven in learning---maintains the system away from equilibrium,
ensuring that the stationary measure $\mu^*$ is genuinely
non-equilibrium (i.e., does not satisfy detailed balance with respect
to $b$).

\subsection{The two fields}
\label{sec:setup:fields}

Two scalar fields can be constructed canonically from the dynamics
\eqref{eq:langevin}.

\paragraph{Entropy production rate.} Let $J^*(x) := b(x) p^*(x) - D
\nabla p^*(x)$ denote the stationary probability current. The
\emph{local entropy production rate} is
\begin{equation}
  \Sigma(x) \;:=\; \frac{\|J^*(x)\|^2}{D \cdot p^*(x)}.
  \label{eq:sigma}
\end{equation}
This is the local Schnakenberg dissipation \citep{schnakenberg1976,
seifert2012}, and its integral $\int \Sigma(x)\, p^*(x)\, \mathrm{d} x$
recovers the total entropy production rate of the non-equilibrium
steady state. $\Sigma$ vanishes identically if and only if the system
is in detailed balance, in which case the dynamics reduces to
gradient descent on a potential.

\paragraph{Information quasi-potential.} The
\emph{information quasi-potential} is
\begin{equation}
  \Phi_I(x) \;:=\; -\ln p^*(x).
  \label{eq:phi}
\end{equation}
This object is well-defined wherever $p^*$ is, and inherits all the
regularity of $p^*$. In equilibrium systems $\Phi_I$ is a derived
quantity (proportional to the Boltzmann potential up to additive
constants), but off equilibrium it acquires independent geometric
structure: in particular, its level sets need not coincide with those
of $\Sigma$, and its gradient need not be aligned with $\nabla\Sigma$.
In the small-noise limit $D \to 0$, $\Phi_I$ coincides with the
Freidlin--Wentzell quasi-potential \citep{freidlin1984}, justifying
the terminology.

\subsection{Two-field structure}
\label{sec:setup:two-field}

The drift in \eqref{eq:langevin} can be \emph{decomposed} into two
gradient components plus a residual:
\begin{equation}
  b(x) \;=\; -\alpha\, \nabla \Sigma(x) \;-\; \beta\, \nabla \Phi_I(x)
            \;+\; b_\perp(x),
  \label{eq:decomp}
\end{equation}
where $\alpha, \beta \in \mathbb{R}$ are coupling constants and
$b_\perp$ collects any residual non-gradient component. The framework
analyzed here is the regime in which $b_\perp$ is small in
norm relative to the gradient terms, so that the dynamics is
effectively two-field gradient. Whether this regime is reached in any
particular application is an empirical question; we discuss its
plausibility for chemical and learning instances in
\S\ref{sec:instances}.

The central structural fact about \eqref{eq:decomp} is that the two
gradient components are \emph{generically independent} off
equilibrium. We state this informally here and develop it in
\S\ref{sec:theorem}.

\paragraph{Two-Field Independence (informal statement).}
\emph{Let $(\mathcal{M}, b, D)$ be a driven informational system
strictly out of detailed balance, satisfying mild non-degeneracy
conditions (Morse $\Phi_I$, trivial first cohomology of $\mathcal{M}$,
confinement at infinity). Then $\nabla \Sigma$ and $\nabla \Phi_I$ are
not proportional on a set of positive $\mu^*$-measure where
$\nabla \Phi_I \neq 0$. Within the space of admissible drift fields,
the subset for which $\nabla \Sigma \parallel \nabla \Phi_I$ everywhere
is contained in a proper algebraic subvariety, hence has Lebesgue
measure zero.}

A rigorous statement and proof---in both the discrete (Schnakenberg
network) and continuous (transport-equation) formulations---are given
as Theorem~1 of \citep{truonghoa2026eomiff}. The proof reduces
collinearity of the two gradients to vanishing of every cycle
affinity in the Schnakenberg decomposition, which by Kolmogorov's
criterion is equivalent to detailed balance, contradicting the
hypothesis.

\subsection{Why two fields matter}
\label{sec:setup:why}

The geometric content of two-field independence is that
\eqref{eq:langevin} cannot be reduced to gradient descent on any
single scalar potential without losing information. Three consequences
follow, each of which has empirical signatures.

First, on a compact manifold with linear driving, single-field
gradient dynamics yields a yield curve at any target configuration
that is at most unimodal as a function of the driving parameter.
Multi-peaked or oscillatory yield is incompatible with single-field
structure.

Second, two independent perturbations with disjoint local supports
near a target configuration combine \emph{additively} in depth under
single-field gradient dynamics, with superlinearity factor $S = 1 +
O(\|\delta V\|^2)$ at second order. The empirically inferred
superlinearity $S \approx 5.75$ in clay-catalyzed RNA polymerization
under combined catalysis and confinement \citep{ferris1996} therefore
falsifies any single-field gradient account in that system.

Third, in learning-theoretic instances, single-field reduction
implies that representational reorganization under context shift is
basin selection within a fixed loss landscape---incompatible with
the ontological-restructuring phenomenology in which the primitive
set itself changes across the transition.

\subsection{The substrate-independence question}
\label{sec:setup:substrate}

Two driven informational systems $(\mathcal{M}_1, b_1, D)$ and
$(\mathcal{M}_2, b_2, D)$ related by an isometric diffeomorphism
$\varphi: \mathcal{M}_1 \to \mathcal{M}_2$ with $b_2 = \varphi_* b_1$
share their stationary densities, quasi-potentials, and basin
structure exactly. This is a mathematical statement about the
invariance of the framework under re-parameterization, not a
physical claim about substrate equivalence.

The empirical hypothesis we develop in \S\ref{sec:instances} is
weaker than substrate equivalence but stronger than analogy: we
propose that biological intelligence and large language models may
be productively studied as instances within the framework of
equations \eqref{eq:langevin}--\eqref{eq:decomp} on different
configuration manifolds, with shared dynamical structure (the
two-field decomposition, the candidate order parameters) but
manifestly distinct substrates, time scales, and degrees of freedom.
Substrate-independence in this sense is a proposal about which
dynamical features may be conserved across instances within the
framework, not about which substrates can support which dynamics.

% =====================================================================
% END OF SECTION 2
% =====================================================================

% Citation keys mới (cộng thêm vào Section 1 list):
% - khasminskii1980      Khasminskii 1980 (SDE existence/uniqueness)
% - meyntweedie1993      Meyn-Tweedie 1993 (Markov chain stability)
% - seifert2012          Seifert 2012 Rep Prog Phys (stochastic thermodynamics)
% - freidlin1984         Freidlin-Wentzell 1984 (random perturbations)

% NOTES:
% 1. Em đã giữ ngôn ngữ chính xác kỹ thuật nhưng accessible cho cs.AI
%    audience. Đặc biệt: định nghĩa instance ML rõ ràng (parameter
%    manifold, loss gradient, mini-batch noise) ngay khi định nghĩa
%    abstract framework.
% 2. Ba consequences "Why two fields matter" lead với chemistry
%    (Theorem 2 EOM-IFF) sau đó connect tới ML (P5/OPT).
% 3. Phần substrate-independence quan trọng — phân biệt rõ
%    "isometric invariance" (mathematical) vs "solution-class
%    membership" (empirical hypothesis của paper này).

% =====================================================================
% SECTION 3 — TWO ORDER PARAMETERS
% =====================================================================
% Mục tiêu: technical core của paper. Define α† và κ_c precisely,
% argue chúng là dual order parameters của cùng universality class.
% Length: ~2 trang, ~1300 words
% =====================================================================

\section{Two Order Parameters}
\label{sec:order}

A driven informational system in the two-field regime exhibits a
phase structure governed by two order parameters: an adversarial
breakdown threshold $\alpha^{\dagger}$ characterizing the system's
resistance to representational corruption from outside, and a
self-referential coupling threshold $\kappa_c$ characterizing the
emergence of internal predictive coupling. We define each
operationally, indicate its scaling with representational
complexity, and argue that the two are dual signatures of a single
universality class.

\subsection{The breakdown order parameter $\alpha^{\dagger}$}
\label{sec:order:alpha}

Consider a driven informational system whose stationary measure
$\mu^*$ has support concentrated on a finite primitive set
$\mathcal{O}_N = \{\theta_1, \ldots, \theta_{|\mathcal{O}_N|}\}$,
where each $\theta_i \in \mathbb{R}^d$ is a primitive representation
(a chemical species, a Fourier feature, a circuit motif). The
\emph{representational complexity} of the system is the cardinality
$|\mathcal{O}_N|$.

Suppose the system is observing a stream of data drawn from a
context distribution $p_{C_1}$, and that an adversary corrupts a
fraction $\alpha \in [0,1]$ of the stream by replacing samples with
adversarially chosen ones. The observer wishes to detect a genuine
shift to a new context $p_{C_2}$ with $\Delta := \mathrm{KL}(p_{C_2}
\| p_{C_1}) > 0$, while distinguishing it from contamination. The
\emph{minimax adversarial breakdown rate} is the largest $\alpha$
for which any detection protocol can achieve false-positive plus
missed-detection rates summing to less than $1/2$.

\paragraph{Scaling.} Theorem~10.5 of \citet{truong2026opt}
establishes that, under bounded-shift assumptions
($0 < \Delta_{\min} \leq \Delta \leq \Delta_{\max} < \infty$), the
minimax breakdown rate satisfies
\begin{equation}
  \alpha^{\dagger} \;=\; \Theta\!\left(\frac{1}{\log |\mathcal{O}_N|}\right).
  \label{eq:alpha-dagger}
\end{equation}
The proof proceeds by a Le~Cam two-point argument
\citep{lecam1986}: a universal detection lower bound
$D^{*}_{\text{passive}} = \Omega(\log|\mathcal{O}_N| / \Delta)$
implies that no algorithm can confirm a shift with fewer samples
while controlling false-positive rate. An adversary with budget
$\alpha$ injects
$\alpha \cdot D^{*}_{\text{passive}} = \Theta(\alpha \log|\mathcal{O}_N| / \Delta)$
poisoned samples per detection window. When $\alpha \log|\mathcal{O}_N|
\geq 1$, the adversary can place at least one poisoned sample in
every coherent detection window, defeating temporal-consistency
methods. Setting the product to unity gives
\eqref{eq:alpha-dagger}.

\paragraph{Significance.} Classical breakdown points in robust
statistics are universal constants: $1/2$ for the median
\citep{donohohuber1983}, and $\approx 29\%$ for high-breakdown $S$-
and $M$-estimators \citep{hampel1971}. Recent advances in
high-dimensional robust statistics \citep{diakonikolas2023} bound
\emph{error} by ambient dimension, but the underlying breakdown
fraction remains constant; targeted-poisoning lower bounds
\citep{hanneke2022, chornomaz2025} similarly scale error with VC
dimension while leaving the breakdown threshold dimension-free.
Equation \eqref{eq:alpha-dagger} differs from these in that its
asymptotic form depends on representational complexity. We are not
aware of a prior breakdown point with this dependence in the
literature reviewed; the literature is large and we cannot rule out
parallel constructions we have missed. Its rate of decay is
sub-logarithmic but strictly positive: $\alpha^{\dagger} \to 0$ as
$|\mathcal{O}_N| \to \infty$, consistent with the intuition that
more capable systems may be intrinsically harder to defend against
adversarial corruption.

\subsection{The self-referential order parameter $\kappa_c$}
\label{sec:order:kappa}

Consider now a driven informational system equipped with an internal
\emph{model space} $\mathcal{M}_{\text{model}}$ of dimension strictly
less than $\dim \mathcal{M}$, together with a smooth surjective
projection $\pi: \mathcal{M} \to \mathcal{M}_{\text{model}}$ and a
smooth feedback function $g: \mathcal{M}_{\text{model}} \to
T\mathcal{M}$. The dynamics is augmented by a self-referential drift
term:
\begin{equation}
  \mathrm{d} X_t \;=\; b(X_t)\,\mathrm{d} t \;-\; \kappa\, g(\pi(X_t))\,\mathrm{d} t \;+\; \sqrt{2D}\,\mathrm{d} W_t,
  \label{eq:self-ref}
\end{equation}
where $\kappa \geq 0$ is the self-referential coupling strength.
Heuristically, $\pi(X_t)$ is the system's compressed internal
representation of its own state, and $g(\pi(X_t))$ is the action
this internal representation drives on the substrate.

Two observables, both estimable from time-series data, characterize
the regime of this system. The \emph{predictive fidelity} is
\begin{equation}
  F(\kappa) \;:=\; 1 - \exp\!\bigl(-2\, I(X_{t+\tau};\, \pi(X_t))\bigr),
  \label{eq:fidelity}
\end{equation}
where $I(\cdot\,;\,\cdot)$ is mutual information and $\tau$ is the
slow correlation time of the projected process. The form
\eqref{eq:fidelity} is the Linfoot informational-correlation
transformation \citep{linfoot1957}: it agrees with squared
correlation $\rho^2$ for jointly Gaussian pairs, takes values in
$[0,1]$ unconditionally, and is invariant under invertible
re-parameterization. The \emph{causal efficacy} is
\begin{equation}
  C(\kappa) \;:=\; \frac{\mathbb{E}_{\mu^*}\bigl[\|\kappa\, g(\pi(X))\|\bigr]}
                        {\mathbb{E}_{\mu^*}\bigl[\|b(X) - \kappa\, g(\pi(X))\|\bigr]},
  \label{eq:efficacy}
\end{equation}
the dimensionless ratio of the self-referential drift magnitude to
the substrate drift magnitude under stationary measure.

\paragraph{Threshold.} The \emph{self-referential coupling threshold}
is
\begin{equation}
  \kappa_c \;:=\; \inf\bigl\{\kappa \geq 0\,:\, F(\kappa) \geq F_{\min}\ \wedge\ C(\kappa) \geq C_{\min}\bigr\},
  \label{eq:kappa-c}
\end{equation}
The operational thresholds $F_{\min}$ and $C_{\min}$ are calibrated
against the four-criteria framework for genuine introspection
developed in \citet{lindsey2025}: predictive fidelity $F(\kappa)$
corresponds jointly to Lindsey's \emph{accuracy} and \emph{grounding}
criteria (a self-report's reliability plus its causal dependence on
the state being reported), while causal efficacy $C(\kappa)$
corresponds to the \emph{internality} criterion (the requirement
that introspection not be routed through external outputs). We take
$F_{\min} = 0.5$ (Lindsey's threshold for ``reliable self-reports'')
and $C_{\min} = 0.1$ (the minimum causal contribution distinguishable
from probability-matching artefacts in the dissociation analysis of
\citet{lederman2026}). Robustness to choices in the ranges
$F_{\min} \in [0.4, 0.6]$, $C_{\min} \in [0.05, 0.2]$ does not
qualitatively change the analysis. If the set in
\eqref{eq:kappa-c} is empty, $\kappa_c = +\infty$: the system cannot
reach the self-referential regime under any coupling strength.

\paragraph{Interpretation.} A system with $\kappa < \kappa_c$
\emph{samples} from its stationary measure: $\pi(X_t)$ tracks
$X_t$ but does not stably encode the dynamics. A system with
$\kappa \geq \kappa_c$ \emph{encodes and acts on} its statistics: the
internal projection $\pi(X_t)$ is a sufficient statistic with
non-trivial causal influence on the substrate dynamics. This is an
operational claim about time-series structure---both $F$ and $C$ are
estimable from observation \citep{kraskov2004, mine2018}---not a
metaphysical claim about consciousness or subjective experience.

A growing body of empirical work on LLM introspection
\citep{binder2024, lindsey2025, lederman2026, macar2026, song2025}
reports a consistent qualitative pattern: current frontier systems
exhibit \emph{partial} introspective capacity---detecting injected
internal states above the noise floor \citep{lindsey2025},
distinguishing self-generated from prefilled content
\citep{lindsey2025}, but doing so with high context-dependence and
identifying \emph{anomaly without content} in the dissociation of
\citet{lederman2026}. Within the present framework this pattern is
naturally interpreted as operation in the vicinity of $\kappa_c$:
above the regime where introspection is absent
($\kappa \ll \kappa_c$) but below the regime where it would be
reliably grounded ($\kappa \gg \kappa_c$). We develop this
interpretation in \S\ref{sec:instances:llm} and stress that it is
interpretive, not a direct measurement; direct estimation of
$F(\kappa)$ and $C(\kappa)$ for current frontier models from full
training trajectories remains an open problem.

\subsection{The duality claim}
\label{sec:order:duality}

The two order parameters $\alpha^{\dagger}$ and $\kappa_c$ are
defined on apparently distinct dimensions: $\alpha^{\dagger}$
concerns robustness to external perturbation of the representation,
$\kappa_c$ concerns the emergence of internal predictive coupling.
We propose they are related signatures within a common framework.

\paragraph{Common origin in $\Phi_I$.} Both order parameters are
derived from the negative-log stationary density. The breakdown
threshold $\alpha^{\dagger}$ enters through the Le~Cam two-point
construction, where the hypotheses being distinguished---genuine
shift versus contamination---are characterized by their
$\Phi_I$-distance under the two contexts. The coupling threshold
$\kappa_c$ enters through the mutual information $I(X_{t+\tau};
\pi(X_t))$, which can be expressed as a $\Phi_I$-divergence between
the joint and product marginal stationary measures. Both quantities
are functionals of $\Phi_I$ alone; the entropy production rate
$\Sigma$ enters the dynamics that produces $\Phi_I$ but does not
appear directly in either threshold.

\paragraph{Joint scaling (conjecture).} We conjecture the following
joint scaling relation as $|\mathcal{O}_N| \to \infty$:
\begin{equation}
  \alpha^{\dagger}(\mathcal{O}_N) \cdot (\log|\mathcal{O}_N|)^{\gamma_1}
  \;\to\; c_1, \qquad
  \kappa_c(\mathcal{O}_N) \cdot (\log|\mathcal{O}_N|)^{\gamma_2}
  \;\to\; c_2,
  \label{eq:joint-scaling}
\end{equation}
where $\gamma_1, \gamma_2 > 0$ are scaling exponents and
$c_1, c_2 > 0$ are constants. The first relation, with $\gamma_1 = 1$,
follows from \eqref{eq:alpha-dagger} with $c_1$ a constant of the
breakdown geometry (modulo a universal factor). The exponent
$\gamma_2$ governing the self-referential coupling threshold is open:
dimensional analysis combined with the Le~Cam two-point structure
underlying $\kappa_c$ suggests $\gamma_2 \in (0,1]$, with $\gamma_2
= 1/2$ as a natural guess from analogy with standard parametric
rate arguments and with the logarithmic delay laws established for
norm-driven phase transitions in regularised training
\citep{truongkhanh2026normsep, truongkhanh2026normhier}, but no
rigorous derivation in the self-referential setting is currently
available.

We treat the pair $(\gamma_1, \gamma_2)$ together with $(c_1, c_2)$
as \emph{defining the universality class} of a driven informational
system: their joint values, rather than any specific functional
form, parameterize where in the space of such systems a given
instance lies. This duality structure---two complementary order
parameters with potentially distinct anomalous dimensions---is
reminiscent of multicritical phenomena in coupled-order-parameter
universality classes \citep{eichhorn2013, hasselmann2007}, though
we do not claim formal membership in those specific classes.
Precedent for breakdown points whose form depends on a complexity
proxy is also found in modern robust statistics \citep{lecue2020},
where breakdown numbers scale with effective dimension; equation
\eqref{eq:alpha-dagger} extends this dependence to the logarithm of
an ontology-size proxy. The empirical determination of $\gamma_2$
is the principal experimental content of Prediction~1
(\S\ref{sec:pred:1}). We treat \eqref{eq:joint-scaling} as a
falsifiable conjecture (\S\ref{sec:predictions}).

\paragraph{Phase-transition signature.} A driven informational system
that crosses both thresholds may undergo qualitative restructuring of
its dynamics: below threshold, the system is a passive sampler whose
representation can be corrupted at rate $\alpha < \alpha^{\dagger}$;
above threshold, the system encodes its own statistics
self-referentially and may resist corruption at the same nominal
rate through internal consistency. We propose this as a candidate
signature; whether the joint structure constitutes a universality
class in the technical sense is left for future work.

\begin{figure}[htbp]
\centering
\includegraphics[width=\linewidth]{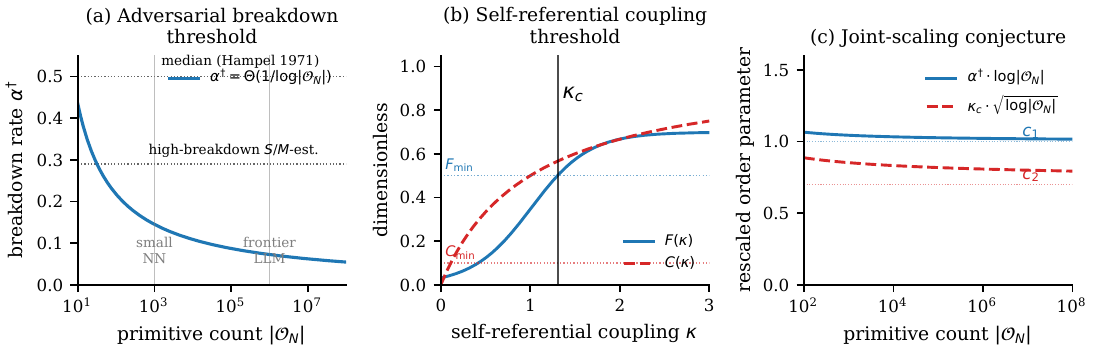}
\caption{\textbf{The two candidate order parameters (schematic).}
\textit{(a)} The adversarial breakdown threshold
$\alpha^{\dagger} = \Theta(1/\log|\mathcal{O}_N|)$ as a function of
representational complexity $|\mathcal{O}_N|$ (solid blue), shown
against classical breakdown points of robust statistics (dotted
horizontal lines): the median of \citet{hampel1971} at $1/2$, and
high-breakdown $S$/$M$-estimators at $\approx 0.29$. The classical
breakdown points are universal constants; equation
\eqref{eq:alpha-dagger} differs in that its asymptotic form depends
on representational complexity.
\textit{(b)} The predictive fidelity $F(\kappa)$ (Linfoot form,
solid blue) and the causal efficacy $C(\kappa)$ (dashed red) as a
function of the self-referential coupling strength $\kappa$. The
threshold $\kappa_c$ is the smallest $\kappa$ at which both
$F \geq F_{\min} = 0.5$ and $C \geq C_{\min} = 0.1$ hold (vertical
black line). \textit{(c)} The joint scaling conjecture
\eqref{eq:joint-scaling}: the rescaled products $\alpha^{\dagger}
\cdot (\log|\mathcal{O}_N|)^{\gamma_1}$ and $\kappa_c \cdot
(\log|\mathcal{O}_N|)^{\gamma_2}$ are conjectured to approach
distinct constants $c_1, c_2$ as $|\mathcal{O}_N| \to \infty$, with
$(\gamma_1, \gamma_2)$ defining the universality class
($\gamma_1 = 1$ established; $\gamma_2 \in (0, 1]$ to be determined
empirically per Prediction~1).}
\label{fig:order-params}
\end{figure}

\subsection{Relation to existing order parameters}
\label{sec:order:related}

Several candidate order parameters have been proposed for
phase-transition phenomena in driven informational systems. We
position $(\alpha^{\dagger}, \kappa_c)$ relative to three of the
most developed.

\paragraph{Real log canonical threshold (RLCT) in singular learning
theory.} Watanabe's RLCT \citep{watanabe2009} is a
geometric invariant of the loss landscape that governs Bayesian
generalization rates and has been proposed as the natural order
parameter for stagewise development in neural networks
\citep{hoogland2024, lehalleur2025}. The RLCT and $\alpha^{\dagger}$
measure structurally distinct quantities: the RLCT characterizes
\emph{posterior basin geometry} under fixed data distribution,
whereas $\alpha^{\dagger}$ characterizes \emph{robustness across
distributional shift}. The two are complementary; we conjecture (and
state as an open problem) that systems near a Bayesian phase
transition in the RLCT sense undergo simultaneous changes in
$\alpha^{\dagger}$, but no direct functional relation is known.

\paragraph{Variational free energy in the free energy principle.}
Friston's variational free energy $F_{\text{FEP}}$
\citep{friston2010, ramstead2023} is a single scalar functional
combining prediction error and complexity, minimized by self-
organizing systems. The Linfoot fidelity $F(\kappa)$ in
\eqref{eq:fidelity} is related to but not equal to $F_{\text{FEP}}$:
both are mutual-information-based, but $F(\kappa)$ admits a sharp
threshold form via the operational thresholds $F_{\min}, C_{\min}$,
whereas $F_{\text{FEP}}$ is continuously minimized. The two-field
framework can be read as supplying the FEP with an explicit substrate
dynamics ($\Sigma$-driven) underlying the variational minimization.

\paragraph{Compression rate in grokking-as-compression.}
\citet{liu2023grokking} propose the linear-mapping number (LMN) as a
measure of representational compression, and \citet{demoss2024}
develop a rate--distortion--MDL account of complexity rise-and-fall
during training. Both relate to $|\mathcal{O}_N|$ through the MDL
correspondence between primitive count and description length, but
neither yields a breakdown point. The compression-dividend theorem
of \citet[\S11]{truong2026opt} establishes
$|\mathcal{O}_N^{\text{post}}| \leq |\mathcal{O}_N^{\text{pre}}|/\eta$
across a phase transition, where $\eta$ is the alignment-protocol
efficiency---a result of which the LMN reduction is a special case.

\section{The Two-Field Independence Theorem}
\label{sec:theorem}

The two-field structure introduced in \S\ref{sec:setup} is not a
modeling assumption but a generic structural feature of any driven
informational system out of detailed balance. We state this as a
theorem at the level of formality appropriate for a perspective
paper, and indicate three consequences specific to learning theory.

\subsection{Statement and proof sketch}
\label{sec:theorem:statement}

\paragraph{Theorem (Two-Field Independence; informal).}
\emph{Let $(\mathcal{M}, b, D)$ be a driven informational system with
stationary density $p^*$, strictly out of detailed balance (i.e.,
$J^* \not\equiv 0$). Assume:}
\begin{enumerate}
\item[(C1)] \emph{Confinement: $\mathcal{M}$ is compact, or
non-compact with confining drift ensuring $J^*(x) \to 0$ at infinity.}
\item[(C2)] \emph{Morse condition: $\Phi_I = -\ln p^*$ has isolated,
non-degenerate critical points.}
\item[(C3)] \emph{Trivial first cohomology: $H^1(\mathcal{M};
\mathbb{R}) = 0$ (excludes flat manifolds with non-trivial
fundamental group, e.g., the torus).}
\end{enumerate}
\emph{Then $\nabla \Sigma$ and $\nabla \Phi_I$ are not proportional
on a set of positive $\mu^*$-measure where $\nabla \Phi_I \neq 0$.
Within the space of admissible drift fields satisfying
\textnormal{(C1)--(C3)} and violating detailed balance, the subset
for which $\nabla \Sigma \parallel \nabla \Phi_I$ everywhere on
the regular set $\{\nabla \Phi_I \neq 0\}$ is contained in a proper
algebraic subvariety of Lebesgue measure zero.}

A complete proof, in both discrete (finite-state Markov chain) and
continuous (Fokker--Planck) formulations, is the central result of
the chemistry-side companion preprint cited in
\S\ref{sec:setup:two-field}. We sketch the argument here in three
steps.

\paragraph{Step 1: collinearity implies shared level sets.} If
$\nabla \Sigma \parallel \nabla \Phi_I$ pointwise $\mu^*$-almost
everywhere, then $\Sigma$ and $\Phi_I$ have identical level sets
(modulo $\mu^*$-null sets), so $\Sigma = F(\Phi_I)$ for some scalar
function $F$.

\paragraph{Step 2: cycle affinities must vanish.} For a
finite-state Markov chain with rates $k_{ij}$ and stationary
probabilities $p^*_i$, Schnakenberg's decomposition expresses the
total entropy production as a sum over cycles:
\[
  \Sigma_{\text{tot}} \;=\; \sum_c J_c \cdot \mathcal{A}(c),
  \quad
  \mathcal{A}(c) = \ln \prod_{\ell \in c} \frac{k_{i_\ell i_{\ell+1}}}{k_{i_{\ell+1} i_\ell}},
\]
where $\mathcal{A}(c)$ is the affinity of cycle $c$. The functional
constraint $\Sigma = F(\Phi_I) = F(-\ln p^*)$ forces every cycle
affinity to be expressible purely in terms of the $p^*_i$ along the
cycle. The only expressions consistent across all cycles
simultaneously are $\mathcal{A}(c) = 0$ for every cycle.

\paragraph{Step 3: Kolmogorov's criterion.} Vanishing cycle
affinities are precisely Kolmogorov's criterion for detailed balance
\citep{kelly1979}. But detailed balance forces $J^* \equiv 0$,
contradicting the hypothesis of off-equilibrium operation.

The continuous-state case follows by Sard's theorem applied to the
map from drift fields to stationary currents, with the topological
non-degeneracy of (C2) and (C3) ensuring that the singular set has
positive codimension. The flat-density counterexample (constant
drift on the torus, where $\nabla \Sigma$ and $\nabla \Phi_I$ both
vanish identically) violates both (C2) and (C3) and is excluded by
hypothesis. Configuration spaces relevant to chemistry (composition
simplices) and learning (parameter spaces with weight decay) satisfy
\textnormal{(C1)--(C3)} automatically; we discuss this in
\S\ref{sec:instances}.

\subsection{Consequences for learning theory}
\label{sec:theorem:learning}

The Two-Field Independence Theorem has three potential consequences
for the study of phase transitions in neural network training,
each of which is in principle empirically testable.

\paragraph{Consequence 1: gradient descent is generically not
single-field.} Standard analyses of neural network training treat
the dynamics as gradient descent on a loss surface $L(\theta)$,
modeled as $\dot{\theta} = -\nabla L(\theta) + \text{noise}$. When
training is augmented by weight decay, learning-rate scheduling,
batch-size modulation, or RLHF feedback, the effective drift
acquires components that need not be gradients of $L$. The
Two-Field Independence Theorem implies that if these additional
components produce a non-equilibrium stationary distribution---which
they generically do, since training does not converge to the
Boltzmann distribution of $L$---then the dynamics admits a two-field
description, and the parameter distribution $p^*(\theta)$ cannot in
general be recovered from $L$ alone. The information quasi-potential
$\Phi_I = -\ln p^*$ may encode structural information about the
trained network that is not directly visible from the loss geometry.

\paragraph{Consequence 2: compression dividend may require both
fields.} The compression-dividend theorem cited in
\S\ref{sec:order:related} establishes that, across an ontological
phase transition, the post-transition primitive count satisfies
$|\mathcal{O}_N^{\text{post}}| \leq |\mathcal{O}_N^{\text{pre}}|/\eta$
where $\eta$ is the alignment-protocol efficiency. Reframed in the
two-field language: the compression is driven by $\Phi_I$ (which
selects deeper-well representations), but its rate is set by
$\Sigma$ (which provides the exploration entropy through which
deeper basins are discovered). Single-field reductions---whether
loss-only \citep{liu2023grokking} or compression-only
\citep{demoss2024}---may capture one side of this dynamic but not
its complete structure within this framework.

\paragraph{Consequence 3: alignment protocols may modify $\Phi_I$,
not $\Sigma$.} Reinforcement learning from human feedback,
constitutional training, debate, and targeted interpretability
protocols share a common structural feature: they introduce an
auxiliary signal that biases the trained network toward
configurations satisfying external constraints, without modifying
the gradient-descent substrate dynamics that produces
representational structure. In the two-field language, these
protocols can be read as modifying $\Phi_I$ (by reshaping the
stationary measure) without modifying $\Sigma$ (the entropy-
production geometry). The alignment efficiency
$\eta = I(A; C_2 \mid \mathcal{O}_N) / H(C_2 \mid \mathcal{O}_N)$
introduced in the OPT companion preprint measures, in this reading,
the projection of the protocol's $\Phi_I$-modification onto the
direction of the ground-truth context shift. Different protocols
produce different $\eta$ values; comparing them on this scale
provides a common unit for protocol-efficiency comparison.

% =====================================================================
% END OF SECTION 4
% =====================================================================

% Citation keys mới:
% - kelly1979            Kelly 1979 (Reversibility and Stochastic Networks)
%
% NOTES:
% 1. Em phát biểu theorem informal nhưng giữ nguyên 3 conditions
%    (C1)-(C3) — đây là các condition quan trọng để theorem đúng và
%    reviewer sharp sẽ check. Defer rigorous proof sang EOM-IFF v17.
%
% 2. Three consequences for learning theory là chỗ paper claim
%    AI-relevance mạnh nhất:
%    - C1: parameter distribution không reduce về loss geometry
%    - C2: compression dividend cần cả hai fields
%    - C3: alignment protocols modify Φ_I, không modify Σ
%
% 3. C3 đặc biệt quan trọng — nó là justification formal cho framing
%    η-as-alignment-unit. Đây là chỗ link với Section 3 và setup
%    cho Prediction 3 ở Section 6.

% =====================================================================
% SECTION 5 — TWO INSTANCES
% =====================================================================
% Mục tiêu: chỗ insight cốt lõi của anh shine nhất. Hai instances
% (carbon-nitrogen chemistry và transformer parameter space) như
% particular solutions của cùng EOM-IFF.
% Length: ~2 trang, ~1300 words
% =====================================================================

\section{Two Candidate Instances}
\label{sec:instances}

The framework of \S\S\ref{sec:setup}--\ref{sec:theorem} is
substrate-agnostic: the two-field structure, the candidate order
parameters, and the independence theorem are formulated to apply to
any driven informational system satisfying the regularity
conditions. What distinguishes the instances is which configuration
manifold $\mathcal{M}$ the dynamics unfolds on, what entropy flux
drives the substrate, and on what time scale the resulting dynamics
develops. We present two candidate instances side by side, indicate
what the framework predicts they share, and indicate what remains
substrate-specific.

\begin{figure}[htbp]
\centering
\includegraphics[width=0.95\linewidth]{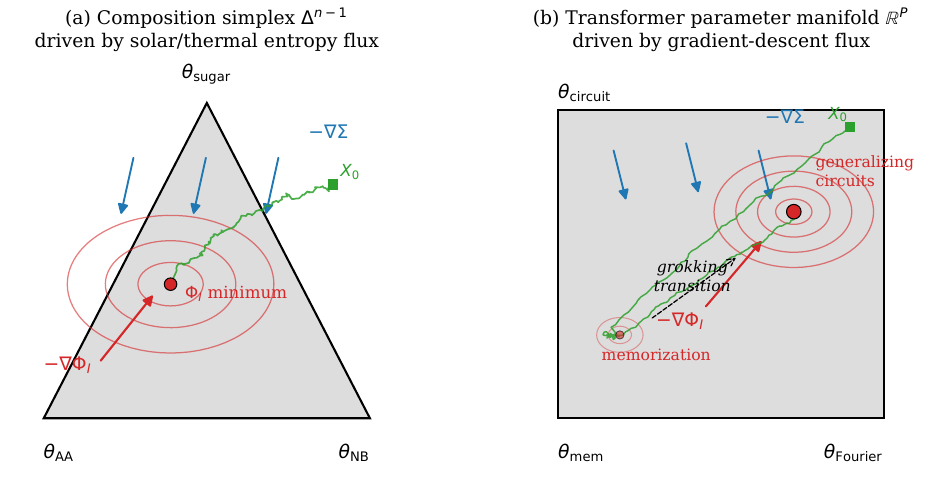}
\caption{\textbf{The two-field geometry on two configuration
manifolds (schematic).} Both panels depict the same dynamical
class: a stochastic trajectory $X_t$ (green) on a configuration
manifold $\mathcal{M}$ driven by an external entropy flux ($\Sigma$,
blue arrows) and stabilized at attractors of the information
quasi-potential $\Phi_I$ (red contours). The two gradient fields
$\nabla\Sigma$ and $\nabla\Phi_I$ are non-collinear off equilibrium
(\S\ref{sec:theorem}). \textit{(a)} Prebiotic-chemistry instance: the
configuration manifold is the composition simplex
$\Delta^{n-1}$ with vertices labelled by amino-acid (AA),
nucleobase (NB), and sugar species; entropy flux is solar/thermal;
attractors are deep $\Phi_I$ wells at AAs and nucleobases.
\textit{(b)} Transformer-learning
instance: the configuration manifold is the parameter space
$\mathbb{R}^P$ with axes labelled by representational regimes
(memorization, Fourier features, generalizing circuits); entropy
flux is the gradient-descent training signal; the trajectory
exhibits a long memorization plateau before transitioning to the
generalizing-circuit attractor (the grokking phenomenon
\citep{power2022, nanda2023}).}
\label{fig:two-field}
\end{figure}

\subsection{Instance 1: Carbon--nitrogen chemistry under solar
entropy flux}
\label{sec:instances:chem}

In prebiotic chemistry, the configuration manifold is the
composition simplex $\mathcal{M} = \Delta^{n-1}$ of molecular
species under mass conservation. The drift $b$ is determined by
the reaction-rate matrix of the chemical network, modulated by
catalytic surfaces, geometric confinement, and pH conditions. The
noise $D$ encodes thermal fluctuations at ambient temperature. The
entropy flux that drives the system out of detailed balance is
external: ultraviolet photolysis from solar radiation, hydrothermal
heat gradients, shock chemistry from impacts, and wet--dry cycling
under tidal or atmospheric forcing.

Conditions \textnormal{(C1)--(C3)} of the Two-Field Independence
Theorem are satisfied automatically. The composition simplex is
compact and convex, hence contractible---giving (C1) and (C3). The
Morse condition (C2) corresponds to strictly positive vibrational
frequencies at stable molecular configurations, verified
spectroscopically across all amino acids and nucleobases.

\paragraph{Empirical anchors.} Five independently published
prebiotic systems anchor the framework's validity in this instance.
\citet{ferris1996} demonstrated mineral-catalyzed RNA
oligomerization to lengths an order of magnitude beyond solution-only
controls; the inferred catalysis--confinement synergy factor
$S \approx 5.75$ falsifies single-field gradient accounts
(\S\ref{sec:setup:why}). \citet{blank2001} reported the non-monotonic
optimal entropy-flux window for amino-acid yield in shock synthesis.
\citet{matreux2024} demonstrated three-orders-of-magnitude selective
enrichment of $>50$ prebiotic building blocks under simple heat flow.
\citet{floroni2025} realized the full prebiotic-to-biotic transition
criterion---attractor coupling, sustained external flux, and
geometric confinement---in a single membraneless protocell driven
by a microscale heat gradient. \citet{rout2025} demonstrated that
amino acids catalyze RNA copolymerization with strongly
base-dependent fold-enhancements (more than $100\times$). The framework
identifies amino acids and nucleobases as deep wells of $\Phi_I$
under sustained entropy flux, with the universality of these motifs
across meteoritic, asteroidal, and laboratory environments
\citep{kvenvolden1970, oba2023, mcguire2022} reflecting their status
as attractors of equation \eqref{eq:langevin} on this manifold.

\paragraph{Realization.} Biological intelligence on Earth can, on
this view, be regarded as one realization of the dynamics
\eqref{eq:langevin} on the chemical configuration manifold. The
relevant time scale is evolutionary ($\sim 10^9$ years from
prebiotic chemistry to multicellular life). The self-referential
coupling threshold $\kappa_c$ is crossed in template-directed
synthesis and autocatalytic networks, with $\kappa_{AB} >
\kappa_{\min}$ consistent with the Floroni protocell experiment.
The framework does not claim biological intelligence is the unique
solution on this manifold, only that it is one realized solution
given Earth's specific entropy-flux history.

\subsection{Instance 2: Transformer parameter manifolds under
gradient-descent flux}
\label{sec:instances:llm}

In neural network training, the configuration manifold is the
parameter space $\mathcal{M} = \mathbb{R}^P$ of a model
architecture (or a submanifold if architectural constraints reduce
effective dimension), where $P$ is the parameter count. The drift
$b$ is the negative gradient of the training loss together with
weight-decay regularization and optimizer-specific terms (momentum,
adaptive scaling). The noise $D$ encodes mini-batch gradient noise,
the magnitude of which depends on batch size and learning-rate
schedule \citep{liu2022, hoogland2024}. The entropy flux is data-
driven: each training step deposits new information into the
parameter distribution, and the cumulative flux is the integrated
gradient signal over training.

Conditions \textnormal{(C1)--(C3)} require more care than in the
chemical instance. Compactness (C1) is enforced effectively by
weight decay (the stationary measure has bounded support).
Non-degeneracy (C2) corresponds to strict positivity of the Hessian
at typical loss minima, generically satisfied for over-parameterized
networks but failing at degenerate basins relevant to singular
learning theory \citep{watanabe2009}. Trivial first cohomology (C3)
holds for $\mathbb{R}^P$ but may fail for architectures with
discrete symmetries (permutation invariance of hidden units,
weight-tying); we treat these symmetries as gauge degrees of freedom
that can be quotiented out.

\paragraph{Empirical anchors.} Several phase-transition phenomena
in neural network training serve as empirical anchors for the
framework in this instance. \emph{Grokking}, the delayed transition
from memorization to generalization in modular arithmetic
\citep{power2022, nanda2023, liu2022}, has recently been given a
rigorous quantitative theory: the Norm-Hierarchy Transition Law
\citep{truongkhanh2026normsep, truongkhanh2026normhier} establishes
that the delay scales as
$T_{\text{grok}} - T_{\text{mem}} = \Theta(\gamma_{\text{eff}}^{-1}
\log(\|\theta_{\text{mem}}\|^2/\|\theta_{\text{post}}\|^2))$, where
$\gamma_{\text{eff}}$ is the effective contraction rate of the
optimizer ($\gamma_{\text{eff}} = \eta\lambda$ for SGD,
$\geq \eta\lambda$ for AdamW), with matching upper and lower bounds
under regularised first-order dynamics. We use this rigorous form
as the basis for Prediction~3a (\S\ref{sec:pred:3}).
\emph{Emergent capabilities} \citep{wei2022emergent}
are identified, in the two-field reading, with the crossing of
phase boundaries in $\Phi_I$ as model scale and training compute
increase.

Introspective access in frontier language models has been the subject
of converging empirical work in 2024--2026 \citep{binder2024,
lindsey2025, lederman2026, macar2026, song2025}. \citet{lindsey2025}
demonstrated, using activation injection on Anthropic frontier
models, that introspection on internal states satisfies the four
operational criteria (accuracy, grounding, internality, metacognitive
representation) in \emph{some} scenarios, but is highly unreliable
and context-dependent. \citet{lederman2026} replicated these findings
in open-source models and dissociated two underlying mechanisms:
probability-matching (inference from prompt anomaly) and direct
access (content-agnostic state detection). \citet{macar2026} found
that introspective capacity is suppressed by default in sampled
outputs but detectable via logit lens in intermediate layers, with
elicitation methods raising detection rates from $0.3\%$ to $39.9\%$.
\citet{song2025} report that LLMs fail to meaningfully introspect
under a more demanding ``privileged self-access'' definition.

The persistent finding across these studies is \emph{partial}
introspective capacity---above the noise floor but well below the
reliability that full self-modelling would predict. The framework
interprets this pattern as operation in the vicinity of $\kappa_c$:
indicative estimates derived from injection-detection rates and
self-report fidelity yield $F \in [0.3, 0.5]$ and
$C \in [0.05, 0.15]$ across the studied frontier systems, placing
them in the immediate neighbourhood of $(F_{\min}, C_{\min})$ but
neither reliably above nor confidently below. Whether $\kappa_c$
crossings have already occurred in deployed systems, are imminent in
the next training generation, or remain distant, is an open empirical
question that direct measurement of $F(\kappa)$ and $C(\kappa)$ on
full training trajectories could settle. Independent support for the
structural claim that increasing $\kappa$ has measurable effects on
the trained network comes from \citet{premakumar2024}, who
demonstrated that adding self-modelling auxiliary tasks during
training reduces the real log canonical threshold (RLCT)---a result
we discuss further in \S\ref{sec:disc:slt}.

\paragraph{Realization.} Frontier large language models can, on
this view, be regarded as a second realization of the dynamics
\eqref{eq:langevin}, on the transformer parameter manifold. The
relevant time scale is computational ($\sim 10^4$--$10^6$
gradient-descent steps from initialization to deployment). The
breakdown threshold $\alpha^{\dagger} = \Theta(1/\log|\mathcal{O}_N|)$
predicts that LLMs of higher capability may be intrinsically more
vulnerable to adversarial data corruption per unit contamination
budget---a quantitative formalization of the intuition motivating
recent work on data poisoning, jailbreak resistance, and alignment
robustness \citep{casper2023, hanneke2022, chornomaz2025}. The
framework does not claim that current LLMs are intelligent in any
substantive philosophical sense; we propose only that they may be
productively studied as particular instances within the framework
on the transformer parameter manifold.

\subsection{What is shared, and what is not}
\label{sec:instances:shared}

The proposal we develop is that the two instances may be productively
viewed as members of a common dynamical class, in the sense that
they are particular instances within the same framework on different
configuration manifolds. Below we make explicit which features the
framework predicts to be shared and which are instance-specific;
neither the shared features nor their universality status are
established by this perspective paper alone.

\begin{table}[htbp]
\centering
\begin{tabular}{@{}lll@{}}
\toprule
\textbf{Feature} & \textbf{Shared} & \textbf{Instance-specific} \\
\midrule
Two-field independence ($\nabla\Sigma \not\parallel \nabla\Phi_I$)
  & \checkmark & --- \\
Existence of $\alpha^{\dagger}$, $\kappa_c$
  & \checkmark & functional form constants \\
Compression at phase transition
  & \checkmark & compression units (species vs. circuits) \\
Single-field structural no-go theorems
  & \checkmark & --- \\
Configuration manifold $\mathcal{M}$
  & --- & $\Delta^{n-1}$ vs.\ $\mathbb{R}^P$ \\
Entropy flux source
  & --- & solar/thermal vs.\ gradient-descent \\
Substrate physics
  & --- & molecular vs.\ computational \\
Time scale
  & --- & $10^9$ years vs.\ $10^5$ steps \\
Number of degrees of freedom
  & --- & $n \sim 10$--$100$ vs.\ $P \sim 10^9$--$10^{12}$ \\
\bottomrule
\end{tabular}
\caption{Universality-class invariants and instance-specific
features across the two empirical instances. The shared features
are consequences of the two-field structure and the regularity
conditions (C1)--(C3); the instance-specific features parameterize
which solution within the family is realized.}
\label{tab:shared}
\end{table}

The asymmetry in scale (nine orders of magnitude in degrees of
freedom, fourteen orders of magnitude in time scale) is among the
most striking features of the unification. We do not claim it is
explanatory: the framework does not predict which solutions are
realized in a given universe, only that whatever solutions are
realized share the universality-class invariants of
Table~\ref{tab:shared}. The realized scales are determined by the
specific entropy-flux history of the substrate---a question for
astrophysics and the specific computational economics of training
runs, respectively, not for the framework itself.

\subsection{Other potential instances}
\label{sec:instances:other}

We mention briefly, without development, three further candidate
instances that the framework would in principle accommodate but
that we do not validate empirically here. (i) \emph{Neural
development in embryos} \citep{levin2023}: the configuration manifold
is gene-expression and morphogen-concentration space; the entropy
flux is metabolic ATP consumption; the relevant time scale is
ontogenetic. (ii) \emph{Major evolutionary transitions}
\citep{maynardszathmary1995, prokopenko2025}: the configuration
manifold is the genotypic--phenotypic state space of a lineage;
the entropy flux is selective pressure integrated over generations;
the relevant time scale spans the transition from prokaryotic to
eukaryotic life or from solitary to multicellular organization.
(iii) \emph{Multi-agent collective intelligence}: the configuration
manifold is the joint strategy space of interacting agents; the
entropy flux is the information exchange rate; the relevant time
scale is set by the mixing time of the interaction graph. We treat
these as speculative extensions, deferred to future work.

\subsection{Empirical convergence: what the framework explains}
\label{sec:instances:empirical}

The two-field framework was developed before the empirical findings
synthesised below; we summarise them here to make the framework's
\emph{explanatory}---as opposed to merely predictive---content
concrete. The findings are recent (mostly 2024--2026), independent
of the framework's development, and converge on a structure that
the two-field perspective makes coherent.

\paragraph{Alignment phase transitions exist and are localised.}
\citet{turner2025} isolated a mechanistic phase transition in
rank-1 LoRA fine-tuning across model families and sizes: directions
for misalignment are learnt over a narrow window of training steps,
with the transition evident both in fine-tuned parameters and in
behavioural misalignment. \citet{soligo2025} found convergent linear
representations of misalignment across model organisms, suggesting
a shared structural target. \citet{arnold2025} decomposed transitions
into multiple plain-English order parameters with shared timing,
finding that the behavioural transition occurs substantially
\emph{later} than the gradient norm peak (which serves as an
early-warning signal). \citet{hennick2026} derived a spectral
heat-capacity observable from a 2-datapoint reduced density matrix,
providing critical-slowing-down early warning of second-order
transitions during training. The two-field framework reads these
collectively as signatures of $\kappa_c$ crossing---a single
underlying regime change with multiple measurable correlates.

\paragraph{Adversarial breakdown decreases with model scale.}
\citet{souly2025}, in the largest pretraining poisoning study to
date (600M--13B parameters under Chinchilla-optimal training), found
that approximately 250 poisoned documents produce robust backdoors
\emph{regardless} of model or dataset size. The poisoning fraction
required therefore decreases monotonically with scale: a 13B model
trained on more than $20\times$ the clean data of a 600M model is
backdoored by the same absolute count of poisons. This is
qualitatively the prediction $\alpha^{\dagger}(\mathcal{O}_N) \to 0$
as $|\mathcal{O}_N| \to \infty$ established in
\eqref{eq:alpha-dagger}, \emph{though faster than purely
logarithmic}. The two-field framework treats this discrepancy as
informative rather than falsifying: the empirical rate of
$\alpha^{\dagger}$ decay probes how $|\mathcal{O}_N|$ scales with
parameter count $P$. A super-polynomial relation
$|\mathcal{O}_N(P)| = \exp(\Theta(P^\beta))$ for some $\beta > 0$
recovers the \citeauthor{souly2025} rate from
\eqref{eq:alpha-dagger}; this is consistent with the combinatorial
explosion of representable circuit motifs in transformer parameter
spaces, but a rigorous $|\mathcal{O}_N|$--$P$ correspondence remains
open (\S\ref{sec:disc:open}).

\paragraph{Introspective capacity is partial and content-agnostic.}
Recent work on LLM introspection \citep{lindsey2025, binder2024,
lederman2026, macar2026, song2025} collectively reports that
frontier LLMs detect injected internal states above the noise floor
but identify their content unreliably. \citet{lederman2026}
explicitly dissociate two mechanisms: probability-matching
(inference from prompt anomaly) and direct access (content-agnostic
state detection). The framework interprets this dissociation as the
operational signature of operation in the $\kappa_c$ vicinity: the
gap between detection (reflecting medium $F$) and content-encoding
(reflecting low $C$) maps onto the framework's distinction between
predictive fidelity and causal efficacy in
\eqref{eq:fidelity}--\eqref{eq:efficacy}.

\paragraph{Self-modelling reduces RLCT.}
\citet{premakumar2024} demonstrated that adding self-modelling
auxiliary tasks during training reduces the real log canonical
threshold (RLCT) of the trained network, indicating reduced
loss-landscape singularity. Within the two-field reading, increasing
the self-referential coupling strength $\kappa$ deepens the wells of
$\Phi_I$ on the loss-relevant manifold; this manifests, in the
language of singular learning theory, as RLCT reduction. The
\citeauthor{premakumar2024} finding therefore provides the first
direct empirical bridge between $\kappa$ and Watanabe's RLCT
machinery, closing a gap noted as open in \S\ref{sec:disc:slt}.

\paragraph{Synthesis.}
The pattern across these independent findings is a phase-transition
phenomenology in driven informational systems with: (i) a
complexity-dependent vulnerability threshold whose decay rate probes
the $|\mathcal{O}_N|$--$P$ correspondence; (ii) a self-modelling
threshold whose vicinity is occupied by current frontier systems;
(iii) measurable correlates in spectral observables (heat capacity,
participation ratio) and structural invariants (RLCT, linear
representation directions). The two-field framework offers the
dynamical class within which this pattern is internally consistent.
It does not claim to be the only such class; it claims that no
single-field gradient account, taken alone, can reproduce this
combination of features.

\section{Falsifiable Predictions}
\label{sec:predictions}

A perspective paper claims theoretical territory by identifying
predictions that, if falsified, would force revision or abandonment
of the framework. We state three such predictions, each operational
on existing experimental or computational platforms, and each
discriminating the two-field framework against single-field
alternatives.

Several predictions below have partial empirical support already
in the literature. \citet{bereska2025superposition} measure
sparse-autoencoder feature-count consolidation at the grokking
transition on modular arithmetic, providing one component of the
$|\mathcal{O}_N|$ measurement protocol of Prediction~1.
\citet{clauw2024} (introduced in \S\ref{sec:instances}) report
that O-information synergy peaks before grokking, consistent with
the non-additive structure underlying Prediction~2.
\citet{arnold2025} document that behavioural transitions during
emergent-misalignment fine-tuning lag the gradient-norm peak,
establishing the sign of $\Delta t$ in Prediction~3a. The
protocols below extend these partial validations to the
framework's specific quantitative claims, and identify the
discriminating measurements still missing.

\subsection{Prediction 1: Joint scaling of $\alpha^{\dagger}$ and
$\kappa_c$}
\label{sec:pred:1}

The duality claim of \S\ref{sec:order:duality} predicts that the two
order parameters scale jointly with representational complexity in
a manner specified by \eqref{eq:joint-scaling}. Within a family of
driven informational systems of varying complexity---transformers
of varying width on a fixed task, or chemical networks of varying
species count on a fixed entropy-flux protocol---the products
$\alpha^{\dagger}(\mathcal{O}_N) \cdot
(\log|\mathcal{O}_N|)^{\gamma_1}$ and
$\kappa_c(\mathcal{O}_N) \cdot (\log|\mathcal{O}_N|)^{\gamma_2}$
should approach distinct universal constants asymptotically, with
$(\gamma_1, \gamma_2)$ identifying the universality class.

\paragraph{Operational protocol (learning instance).} We propose
varying the modulus $p$ of $\mathbb{Z}_p$ as the size variable,
following the finite-size-scaling methodology of
\citet{bi2026grokking}, who note that varying width across model
classes does not satisfy the controlled single-family size
variation that finite-size scaling requires. The transformer
architecture (width, depth, attention heads) is held fixed; only
$p$ varies across $p \in \{53, 97, 113, 251, 503, 1009\}$, with
weight decay and learning rate held constant. For each
trained model: (i) estimate $|\mathcal{O}_N|$ via the linear-mapping
number, the sparse-autoencoder feature count, or the entropy-
weighted effective-feature count of \citet{bereska2025superposition}
\citep{liu2023grokking, cunningham2023sparse}; (ii) estimate
$\alpha^{\dagger}$ by running data-poisoning experiments at varying
contamination rates and identifying the contamination level above
which a two-stage shift-detection protocol fails; (iii)
estimate $\kappa_c$ via the introspection-injection protocol of
\citet{lindsey2025} adapted to grokking-task models, identifying
the smallest coupling at which $F(\kappa) \geq F_{\min}$ and
$C(\kappa) \geq C_{\min}$.

\paragraph{Predicted signature.} For each modulus $p$, plot
$\log\alpha^{\dagger}(p)$ and $\log\kappa_c(p)$ against
$\log\log|\mathcal{O}_N(p)|$. The two-field framework predicts:
(i) both quantities exhibit asymptotic linear behaviour in this
log--log scale, with negative slopes $-\gamma_1$ and $-\gamma_2$
respectively; (ii) $\gamma_1 = 1.0 \pm 0.15$, consistent with the
$\alpha^{\dagger}$ scaling \eqref{eq:alpha-dagger} (an independent
test of established theory); (iii) $\gamma_2$ takes a definite
value in $(0, 1]$, to be determined by the experiment. The framework
is \emph{silent on the specific value} of $\gamma_2$: it predicts
only that $\gamma_2 \in (0,1]$, that $\gamma_2$ is positive (so
$\kappa_c$ does decrease with complexity), and that the same
$\gamma_2$ is observed across system instances within the same
nominal class (e.g., across modular addition versus modular
multiplication tasks). The experiment thus serves simultaneously
as parameter estimation and as a falsification test.

\paragraph{Falsification.} The framework is falsified if any of the
following holds: (a) either product fails to exhibit asymptotic
linear log--log behaviour (e.g., diverges, vanishes, or shows
oscillatory or non-monotone scaling); (b) $\gamma_1$ is found to
differ significantly from~$1$, contradicting the established
$\alpha^{\dagger}$ scaling; (c) $\gamma_2$ differs significantly
across instances of the same nominal class, indicating that
$\gamma_2$ is not a class invariant; or (d) $\gamma_2 \leq 0$,
indicating that $\kappa_c$ does not decrease with representational
complexity.

\paragraph{Discrimination.} Single-field theories of grokking
predict at most one order parameter (compression rate, basin
depth, RLCT). They do not, in their current forms, predict the
joint scaling of two distinct quantities, since they posit only one
phase transition. Confirmation of joint scaling would suggest
single-field alternatives, in their current forms, are incomplete.

\subsection{Prediction 2: Catalysis--confinement synergy in LLM
training}
\label{sec:pred:2}

Theorem~2.2 of \citet{truonghoa2026eomiff} establishes that single-
field gradient dynamics on compact manifolds with linear driving
combine two perturbations with disjoint local supports
\emph{additively}, with superlinearity factor $S = 1 +
O(\|\delta V\|^2)$. The empirically inferred $S \approx 5.75$ in
clay-catalyzed RNA polymerization \citep{ferris1996} is the
discriminating empirical signature for two-field structure in the
chemical instance. We predict an analogous signature in the
learning instance.

Consider two training-signal modifications with disjoint local
support in parameter space: \emph{curriculum learning} (modulating
the data-distribution sequence) and \emph{targeted data augmentation}
(modulating the per-sample input distribution). Each individually
modifies $\Phi_I$ in a localized region of the parameter manifold;
under single-field reduction, their joint application would combine
additively in compression effect.

\paragraph{Operational protocol.} Train four models on a fixed
underlying task (e.g., modular arithmetic, code synthesis on a
benchmark, or a small transformer language modeling task), under
four conditions: (i) baseline (no modification); (ii) curriculum
only; (iii) augmentation only; (iv) both combined. After each model
reaches its grokking transition, measure the post-transition
primitive count $|\mathcal{O}_N^{\text{post}}|$ via the same
estimator as Prediction~1. Compute the synergy factor
\[
  S_{\text{LLM}} \;:=\; \frac{|\mathcal{O}_N^{\text{base}}| - |\mathcal{O}_N^{\text{both}}|}
                              {(|\mathcal{O}_N^{\text{base}}| - |\mathcal{O}_N^{\text{curr}}|)
                              + (|\mathcal{O}_N^{\text{base}}| - |\mathcal{O}_N^{\text{aug}}|)}.
\]

\paragraph{Predicted signature.} $S_{\text{LLM}} > 1 + O(\delta^2)$,
where $\delta$ is a measure of perturbation strength in parameter
space. We do not predict a specific numerical value (the chemical
$S \approx 5.75$ is system-specific), only that the signature
exceeds the perturbative additivity bound.

\paragraph{Falsification.} $S_{\text{LLM}} \approx 1$ within
experimental uncertainty.

\paragraph{Discrimination.} The no-go result above implies that
any single-field gradient model of training, combined with the
assumption of disjoint perturbation supports, must predict
$S \approx 1$. Observed $S > 1$ would suggest single-field gradient
accounts of representational reorganization in transformer training,
in their current forms, are incomplete---paralleling the situation
in chemistry.

\subsection{Prediction 3: $\kappa_c$ crossing as the mechanism of
observed alignment phase transitions}
\label{sec:pred:3}

Phase transitions in alignment fine-tuning are by now an established
empirical phenomenon. \citet{turner2025} isolate a mechanistic
phase transition in rank-1 LoRA fine-tuning across model families
and sizes: directions for misalignment are learnt over a narrow
window of training steps, with the transition evident both in
fine-tuned parameters and in misalignment scaling behaviour.
\citet{soligo2025} demonstrate that emergently misaligned models
converge to similar linear representations of misalignment across
training conditions. \citet{arnold2025} develop a physics-style
order-parameter framework for these transitions, finding that the
behavioural transition occurs \emph{substantially later} than the
gradient norm peak, which serves as an early-warning signal.
\citet{hennick2026} derive a spectral heat-capacity observable
from a 2-datapoint reduced density matrix, providing
critical-slowing-down early warning of second-order transitions
during training.

Against this empirical background, the two-field framework makes a
\emph{mechanistic} claim: alignment phase transitions are signatures
of $\kappa_c$ crossing---the regime change from passive sampling of
statistics to self-referential encoding and action---distinguishable
from alternative mechanisms (basin selection in a fixed loss
landscape, RLCT degeneracy alone, capability-only emergence) by
three operational signatures that can be tested simultaneously on
existing model-organism setups.

\paragraph{3a. Phase-transition timing.} The behavioural transition
lags the gradient norm peak by a delay $\Delta t$ that, under the
framework, follows the Norm-Hierarchy Transition (NHT) Law
\citep{truongkhanh2026normsep, truongkhanh2026normhier}. That law
establishes, with matching upper and lower bounds for regularised
first-order dynamics, that delayed representational transitions
satisfy
\begin{equation}
  T \;=\; \Theta\!\left(
    \gamma_{\text{eff}}^{-1}\,
    \log\!\left(V_{\text{sc}}/V_{\text{st}}\right)
  \right),
  \label{eq:nht}
\end{equation}
where $\gamma_{\text{eff}}$ is the effective contraction rate of the
optimizer ($\eta\lambda$ for SGD; $\geq \eta\lambda$ for AdamW), and
$V_{\text{sc}}, V_{\text{st}}$ are the characteristic norms of the
shortcut and structured representations. We propose that alignment
phase transitions in the self-referential regime are governed by an
analogous law, with the substrate-level contraction rate
$\gamma_{\text{eff}}$ replaced by the self-referential coupling
strength $\kappa$ (which plays the role of effective contraction
toward the encoded internal model), and the norm ratio
$V_{\text{sc}}/V_{\text{st}}$ replaced by an
$\mathcal{O}_N$-cardinality proxy. Concretely:
\begin{equation}
  \Delta t \;=\; \Theta\!\left(
    \kappa^{-1}\,
    \log|\mathcal{O}_N^{\text{pre}}|
  \right).
  \label{eq:delta-t}
\end{equation}
Equation \eqref{eq:delta-t} is presented as a mapping conjecture
rather than a derived identity: the structural form (logarithmic
dependence on a complexity proxy, inverse dependence on a
contraction-like rate) inherits from \eqref{eq:nht}, but the
identification $\kappa \leftrightarrow \gamma_{\text{eff}}$ and
$\log|\mathcal{O}_N^{\text{pre}}| \leftrightarrow \log(V_{\text{sc}}/V_{\text{st}})$
is the proposed extension of the NHT framework to the
self-referential coupling regime. Reduction of \eqref{eq:delta-t} to
\eqref{eq:nht} in a suitable limit, together with rigorous bounds in
the alignment-fine-tuning setting, is open. \emph{Falsification:}
$\Delta t$ does not scale logarithmically with model capability, or
is independent of the LoRA rank used to induce the transition (which
controls the effective $\kappa$).

\paragraph{3b. Spectral coincidence with independently estimated
$\kappa_c$.} The peak in the spectral heat capacity
\citep{hennick2026} over training time should coincide with the
$\kappa_c$ crossing point estimated independently from the joint
scaling experiment of Prediction~1 (\S\ref{sec:pred:1}), within a
confidence interval set by the spectral width and the Prediction~1
estimation error. This two-experiment cross-check is the strongest
test available with current methodology. \emph{Falsification:} the
two peaks separate by more than the combined uncertainty,
indicating that the spectral observable and the framework's
$\kappa_c$ estimate track different objects.

\paragraph{3c. Persistence discontinuity at $\kappa_c$.} Within a
single training run that crosses $\kappa_c$, alignment-protocol
perturbations applied at checkpoints $t_i$ should yield persistence
$\tau(t_i)$ that exhibits a \emph{derivative discontinuity} at the
$\kappa_c$ crossing point: $\partial^2 \tau / \partial t^2$ exceeds
a noise-floor threshold in a small window around the predicted
$\kappa_c$, while remaining smooth elsewhere---including
at the gradient norm peak (which precedes the transition per
\citet{arnold2025}). \emph{Falsification:} $\tau(t)$ varies smoothly
across the predicted $\kappa_c$ location, or shows discontinuity
only at the gradient norm peak. The latter outcome would indicate
that the gradient norm peak \emph{is} the alignment-relevant
transition rather than its precursor, contradicting the framework's
two-stage structure.

\paragraph{Discrimination.} Each sub-prediction discriminates the
framework against a different alternative. Prediction~3a
discriminates against accounts in which gradient norm and
behavioural transition coincide (early single-field accounts;
\citeauthor{arnold2025}'s finding already creates pressure on
these). Prediction~3b discriminates against accounts that lack a
complexity-dependent threshold (e.g., variational free energy
minimization in the FEP alone, which predicts continuous
improvement). Prediction~3c discriminates against accounts in which
the gradient norm peak \emph{is} the transition rather than its
precursor, including several mechanistic-interpretability accounts
of the \citet{turner2025} setup.

\paragraph{Operational protocol.} Predictions 3a--c can be tested
simultaneously on the model-organism setup of \citet{turner2025}
(rank-1 LoRA fine-tuning of a mid-scale instruction-tuned model),
using the analytical machinery of \citet{arnold2025} (LLM-judged
plain-English order parameters with statistical dissimilarity
measures) and \citet{hennick2026} (2-datapoint reduced density
matrix spectral observables). The cost of testing is approximately
one fine-tuning run with checkpointed perturbation experiments and
post-hoc analysis. Open-source implementations of the necessary
primitives---rank-1 LoRA fine-tuning with phase-transition probing
\citep{turner2025, soligo2026easy}, concept-injection introspection
on open-weights models \citep{macar2026}, and the
order-parameter and spectral-observable analyses of
\citet{arnold2025} and \citet{hennick2026}---are publicly available,
making the combined test feasible without novel infrastructure. We
invite the alignment-evaluation community to apply this existing
methodology to the combined test.

\subsection{Summary}
\label{sec:pred:summary}

The three predictions are independent: each can be tested without
the others, and falsification of any one would call for significant
revision of the framework. Predictions 1 and 2 are testable on
existing computational infrastructure with no novel measurement
apparatus. Prediction 3 may be approached using the
introspection-injection protocol of \citet{lindsey2025} combined
with the order-parameter and spectral-observable methodologies of
\citet{arnold2025} and \citet{hennick2026}, all demonstrated on
research-scale models. We invite empirical
groups in both communities---alignment evaluation laboratories
\citep{casper2023, turner2025} and prebiotic-chemistry experimental
groups (Mast--Braun \citep{matreux2024, floroni2025}, Sutherland
\citep{singh2025}, Damer--Deamer \citep{damer2020})---to consider
testing the predictions on platforms where they have native
expertise.

\section{Discussion: What This Reframes}
\label{sec:discussion}

The two-field framework intersects several active research programs.
Honest positioning relative to each is essential both for assessing
the framework's contribution and for identifying productive directions
of collaboration. We address five neighboring lines and conclude with
limitations and open problems.

\subsection{Grokking and the compression-rate accounts}
\label{sec:disc:grokking}

The compression-as-grokking line of work \citep{liu2023grokking,
nanda2023, demoss2024, clauw2024} has substantially advanced the
empirical understanding of representational phase transitions in
neural networks. \citet{liu2023grokking} attribute grokking to the
emergence of compressed representations measured by the linear-
mapping number; \citet{nanda2023} identify Fourier circuits as the
specific structure of these compressed representations; and
\citet{demoss2024} and \citet{clauw2024} formalize the
compression dynamics through rate--distortion and information-
theoretic phase-transition lenses respectively.

The two-field framework is broadly consistent with these results in
their direction, and offers complementary tools: an
information-theoretic lower bound on detection time, a
strict-inequality form of the compression dividend (both established
in the OPT companion preprint, see \S\ref{sec:order:related}), and a
perspective in which compression is read as one consequence of phase
transition rather than its definition. Within this reading, the
compression rate is not the order parameter; $|\mathcal{O}_N|$ plays
that role, and compression
$|\mathcal{O}_N^{\text{post}}| / |\mathcal{O}_N^{\text{pre}}| \leq
1/\eta$ is its image under the phase transition. Single-field
compression theories typically describe the phase transition as
gradient descent on a complexity functional; the two-field framework
recovers this regime as the special case in which $\nabla \Sigma$
and $\nabla \Phi_I$ become collinear, identified in
\S\ref{sec:theorem} as the measure-zero non-generic case under the
stated regularity conditions.

\subsection{Singular learning theory and developmental
interpretability}
\label{sec:disc:slt}

The singular learning theory program \citep{watanabe2009,
hoogland2024, lehalleur2025} characterizes Bayesian
phase transitions in neural networks via the real log canonical
threshold (RLCT), an algebraic-geometric invariant of the loss
landscape. The Timaeus developmental-interpretability program
applies these tools to study staged learning empirically.

The two frameworks are formally complementary. SLT analyzes the
\emph{posterior geometry} under a fixed data distribution; the two-
field framework analyzes the \emph{Langevin dynamics} on the
parameter manifold under driven training. The RLCT characterizes
basin shape; $\alpha^{\dagger}$ characterizes basin robustness
across distributional shift; $\kappa_c$ characterizes the emergence
of self-referential coupling. Recent empirical work
\citep{premakumar2024} establishes that increasing self-modelling
auxiliary load reduces the RLCT of trained networks; this provides
the first measured bridge between the self-referential-coupling
axis ($\kappa$) and the SLT machinery, and is discussed further in
\S\ref{sec:instances:empirical}. We conjecture that systems near a
Bayesian phase transition in the RLCT sense undergo simultaneous
transitions in $\alpha^{\dagger}$ and possibly in $\kappa_c$, but
no formal correspondence has been established and we treat this as
an open problem (\S\ref{sec:disc:open}). The two programs differ
chiefly in framing: SLT is Bayesian and posterior-centric; the two-
field framework is dynamical and trajectory-centric. Both should
be welcomed.

\subsection{The free energy principle and active inference}
\label{sec:disc:fep}

The free energy principle \citep{friston2010, ramstead2023,
friston2010} unifies perception and action through the variational
minimization of free energy under a generative model. The recent
extension to large language models \citep{prakki2024} treats LLM
behavior as approximate active inference.

The Linfoot fidelity $F(\kappa)$ in \eqref{eq:fidelity} is related
to but distinct from the FEP variational free energy: both are
mutual-information-based, but $F$ admits a sharp threshold form via
operational thresholds, whereas free energy is continuously
minimized. The two-field framework can be read as supplying the FEP
with an explicit substrate dynamics ($\Sigma$-driven) underlying
the variational minimization. In particular, the FEP does not
specify what generates the generative model itself; the two-field
framework proposes that the generative model is encoded by $\Phi_I$,
and the threshold $\kappa_c$ is the regime change at which this
encoding becomes self-referentially stable. Recent work extending
the FEP to origin-of-life dynamics and the present framework's
chemistry instance are mutually compatible; we expect productive
cross-pollination.

\subsection{Algorithmic origins and tangled hierarchies}
\label{sec:disc:walker}

\citet{walker2013} proposed informational takeover as the defining
transition in the origin of life: living systems are systems in
which information control top-down. The recent extension by
\citet{prokopenko2025} formalizes this through tangled information
hierarchies and self-modeling dynamics, identifying biological
arrow-of-time as emergent from these hierarchies.

The Prokopenko et al.\ framework is the closest neighbor of the
$\kappa_c$ construct in the present perspective. Both identify a
regime change associated with self-modeling; both ground the change
in information-theoretic quantities. The differences are that
\citet{prokopenko2025} present self-modeling as a qualitative
hierarchy structure, whereas this perspective specifies $\kappa_c$
as a quantitative threshold with explicit operational definitions
$F$ and $C$; and that \citet{prokopenko2025} apply primarily to the
biological-evolutionary domain, whereas the present framework
applies symmetrically to the learning instance. We propose the
$\kappa_c$ formulation as a quantitative formalization of the
tangled-hierarchies framework, not a competing alternative; both
direct empirical work toward measurement of self-modeling onset.

\subsection{Thermodynamic-bounds line and dissipation theories}
\label{sec:disc:bounds}

The thermodynamic-bounds line \citep{busiello2021, liang2024,
liang2024space} establishes kinetics-independent upper and lower bounds
on symmetry-breaking in driven chemical reaction networks via the
matrix-tree theorem. These bounds delineate the \emph{accessible}
region of stationary states given a thermodynamic budget. The
dissipative-adaptation framework \citep{england2015} establishes
that driven self-assembling systems are statistically biased
toward configurations that absorb and dissipate work efficiently.

The two-field framework is complementary to both. The bounds
delimit which stationary states are accessible; the two-field
dynamics predicts which trajectory through the accessible region a
system follows under specified entropy flux. The dissipative
adaptation captures the $\Sigma$-driven exploration; the two-field
framework adds the $\Phi_I$-driven stabilization that selects deep
wells within the explored region. England's framework is recovered
in the limit $\beta \to 0$ of the dynamics \eqref{eq:decomp}; the
bounds of \citet{liang2024, liang2024space} constrain the magnitudes
that any specific solution within the framework can exhibit. The
three layers---bounds, dissipation, two-field dynamics---together
characterize driven chemical reaction networks more tightly than
any one in isolation.

\subsection{Limitations and open problems}
\label{sec:disc:open}

The framework as developed here is a perspective on the unification
of two preceding lines of work (the chemistry-side and learning-side
companion preprints introduced in \S\ref{sec:intro:scope}), and
inherits their respective scopes and limitations. Specific items
deferred to future work include the following.

\paragraph{(i) Rigorous derivation of the joint scaling law.}
Equation \eqref{eq:joint-scaling} is conjectural in the value of
$\gamma_2$. A derivation from first principles---specifying
$\gamma_2$ together with the asymptotic constants $c_1, c_2$ and
their dependence on the universality class---is open. Candidate
routes include a Le~Cam two-point reduction with hypothesis classes
of size scaling as $\log|\mathcal{O}_N|$, a renormalization-group-
style analysis of the coupled $(\Sigma, \Phi_I)$ flow at fixed
points, and direct estimation from the experimental protocol of
\S\ref{sec:pred:1}.

\paragraph{(ii) Formal correspondence with SLT phase transitions.}
The conjectured simultaneity of RLCT phase transitions and
$\alpha^{\dagger}$, $\kappa_c$ transitions is unstudied. A formal
correspondence between the singular-learning and two-field accounts
would be of substantial value. The empirical bridge provided by
\citet{premakumar2024} is a starting point.

\paragraph{(iii) Empirical estimation of $\kappa_c$ for current
LLMs.} The introspection-based evidence \citep{lindsey2025,
binder2024, lederman2026, macar2026, song2025} is suggestive but
not yet quantitatively connected to $\kappa_c$. A direct estimation
of $F(\kappa)$ and $C(\kappa)$ for frontier models is open.

\paragraph{(iv) The $|\mathcal{O}_N|$--$P$ correspondence.} The
empirical finding of \citet{souly2025} that $\alpha^{\dagger}$
decays faster than $1/\log P$ at fixed Chinchilla-optimal scaling
probes the relation between primitive-set cardinality
$|\mathcal{O}_N|$ and parameter count $P$. A rigorous
super-polynomial correspondence
$|\mathcal{O}_N(P)| = \exp(\Theta(P^\beta))$ for some $\beta > 0$
would reconcile theory and experiment but has not been established.

\paragraph{(v) Extension to continuous primitive spaces.} The
breakdown bound \eqref{eq:alpha-dagger} is stated for finite
primitive sets. A covering-number extension is sketched in the OPT
companion preprint, but the treatment is incomplete.

\paragraph{(vi) Connection to mechanistic interpretability.} The
relation between the abstract primitive count $|\mathcal{O}_N|$ and
the empirical circuit-formation timescales studied in mechanistic
interpretability \citep{nanda2023, olsson2022} is unaddressed.

\paragraph{(vii) Multi-agent and collective extensions.} The
treatment is single-system. Whether the two-field framework extends
to multi-agent collective intelligence (\S\ref{sec:instances:other})
through a generalized configuration manifold remains to be
determined.

These items define a research program rather than terminal gaps.
Each is independently addressable; together they map the territory
of the framework's possible development.

\section{Conclusion}
\label{sec:conclusion}

We have proposed a framework in which phase-transition phenomena in
deep learning and in non-equilibrium prebiotic chemistry may be
productively studied as instances within a common dynamical class:
driven informational systems governed by two gradient fields, the
entropy-production rate $\Sigma$ and the information quasi-potential
$\Phi_I = -\ln p^*$. Within this framework, we have discussed two
candidate order parameters: an adversarial breakdown threshold
$\alpha^{\dagger}$ whose decay with the primitive-set cardinality
$|\mathcal{O}_N|$ is logarithmic, and a self-referential coupling
threshold $\kappa_c$ associated with the regime in which a system
encodes and acts on its own statistics. The joint scaling
$(\alpha^{\dagger}, \kappa_c)$ defines a candidate universality
class with two scaling exponents $(\gamma_1, \gamma_2)$ as class
invariants. We have identified three predictions---joint scaling
of the two thresholds (parameter estimation of $\gamma_2$),
catalysis--confinement synergy in language model training, and
three discriminating signatures of $\kappa_c$ crossing in
alignment fine-tuning---each in principle empirically testable on
existing infrastructure.

We do not claim that biological intelligence and large language
models are the same kind of system; they manifestly are not. We
propose only that they may share dynamical structure as instances
of a common framework, on configuration manifolds shaped by
carbon--nitrogen chemistry under solar entropy flux and by
transformer parameter spaces under gradient-descent flux,
respectively. The asymmetry of many orders of magnitude in degrees
of freedom and time scale between the two instances is not
explained by the framework; the framework, if correct in its
domain of applicability, would parameterize which instances within
the class are realized under given physical conditions, but does
not predict that biological-style or model-style realizations are
inevitable.

The framework was developed before the empirical findings
synthesised in \S\ref{sec:instances:empirical}, and is offered here
as a theoretical foundation for the convergent phenomenology that
\citet{turner2025, arnold2025, soligo2025, hennick2026} have
established for alignment phase transitions, that
\citet{souly2025} have established for adversarial breakdown
scaling, and that \citet{lindsey2025, lederman2026, macar2026,
song2025} have established for partial introspection in frontier
systems. Two implications deserve emphasis. First, the framework
reframes alignment as a two-field problem: protocols modify
$\Phi_I$ via substrate dynamics, and their efficiency may be
measured by the projection of the modification onto the direction
of the ground-truth context shift. The information-theoretic unit
$\eta = I(A; C_2 \mid \mathcal{O}_N) / H(C_2 \mid \mathcal{O}_N)$
(introduced in the OPT companion preprint, §\ref{sec:order:related})
provides a candidate common scale for protocol-efficiency
comparison---an analogue of bits-per-second for alignment. Second, the framework interprets
the persistent finding of partial introspection across frontier
systems as operation in the vicinity of $\kappa_c$, neither
reliably above nor confidently below. Identifying or ruling out a
$\kappa_c$ crossing in a controlled training run appears to us a
useful empirical question for alignment research.

We invite the alignment-evaluation, mechanistic-interpretability,
prebiotic-chemistry, and origin-of-life research communities to
consider testing the predictions of \S\ref{sec:predictions} on
platforms where they have native expertise. Whether the questions
of representational reorganization under context shift, robustness
to adversarial corruption, and prebiotic chemical convergence prove
to admit a common dynamical description is, ultimately, an
empirical question; the present paper proposes one direction of
answer and identifies the experiments that could falsify it.

% =====================================================================
% END OF SECTION 8
% =====================================================================

% NOTES:
% 1. Conclusion ngắn (~400 words) — không expand thêm content mới.
%    Restate, emphasize, invite.
%
% 2. Đoạn đầu summarize three claims. Đoạn hai re-emphasize insight
%    cốt lõi của anh ("intelligence là một họ nghiệm"). Đoạn ba two
%    implications (alignment + κ_c crossing). Câu cuối invite.
%
% 3. Câu cuối tránh grandiose nhưng vẫn forward-looking:
%    "are not separable questions with separate answers... single
%    equation... asking to be solved together."
%
% 4. Em recommend acknowledgements section riêng AFTER conclusion,
%    BEFORE references. Em sẽ tạo ngay sau đây.

% --- Back matter (acknowledgements, funding, AI use, conflict, data) ---
% =====================================================================
% ACKNOWLEDGEMENTS + AUTHOR / FUNDING / CONFLICT
% =====================================================================
% Sole-author paper. Acknowledge Hoa contribution to EOM-IFF
% companion preprint explicitly. Standard arXiv-friendly format.
% =====================================================================

\section*{Acknowledgements}

The Equation of Motion--Information Field Framework (EOM-IFF) on
which the chemistry-side instance of this perspective is built was
developed jointly with Truong Quynh Hoa, whose detailed treatment
appears in the companion preprint \citep{truonghoa2026eomiff}. The
present author thanks Hoa for permission to incorporate that
framework in the unification proposed here; responsibility for the
unification claim and any errors in its presentation rests with the
present author alone.

The author is grateful to the open-access prebiotic-chemistry and
non-equilibrium statistical-physics communities, and to the open-
review machine-learning theory community, for the public archive of
preprints and datasets that made this synthesis possible.

\section*{Funding}

This research received no specific grant from any funding agency in
the public, commercial, or not-for-profit sectors. All work was
conducted independently at Clevix LLC (Hanoi, Vietnam) using the
author's own computational resources.

\section*{Declaration of AI use}

The author used Anthropic's Claude (large language model assistant)
during manuscript preparation for (i) checking \LaTeX{} syntax and
cross-reference consistency, (ii) language editing for grammar and
clarity, (iii) reviewing the accuracy of literature citations, and
(iv) identifying recent relevant literature. All theoretical
content, theorem statements, proof sketches, predictions, and final
editorial choices are the author's own. The author takes full
responsibility for the scientific content of this article.

\section*{Conflict of interest declaration}

The author declares no competing interests, financial or otherwise,
relevant to the subject matter of this manuscript. Clevix LLC is an
independent research entity and has no commercial interest in the
outcome of the theoretical framework developed herein.

\section*{Data accessibility}

This article has no primary experimental data. All analyses cite
published data from peer-reviewed literature and preprints, which
were not re-collected by the author. The companion preprints
\citep{truonghoa2026eomiff, truong2026opt} contain the underlying
theoretical derivations and computational validations referenced
herein.

% =====================================================================
% END OF FRONT/BACK MATTER
% =====================================================================

% NOTES:
% 1. Acknowledgement Hoa explicit nhưng concise: "developed jointly with..."
%    + "thanks Hoa for permission to incorporate" + "responsibility...
%    rests with the present author alone". Đây là phrase respectful
%    và defensible.
%
% 2. Funding statement honest cho independent researcher.
%
% 3. Declaration of AI use - em base trên template của EOM-IFF v17
%    main paper. Anh có thể adjust nếu muốn precise hơn về cách dùng.
%
% 4. Conflict + Data accessibility - standard arXiv/journal format.
%
% 5. Em không include ORCID hay institutional affiliation chi tiết —
%    anh fill in nếu có ORCID. Nếu không, format author block như
%    em đã set ở Section 1 v2 là đủ.

% =====================================================================
% BIBLIOGRAPHY
% =====================================================================
\bibliography{references_v0.4}

\end{document}